\documentclass[letterpaper, 10 pt, conference]{ieeeconf}  
\IEEEoverridecommandlockouts                  
\usepackage[letterpaper, left=0.75in, right=0.75in, bottom=0.75in, top=1in]{geometry}
\usepackage{graphics} 
\usepackage{amsmath}  
\usepackage{amssymb}  
\usepackage[export]{adjustbox} 
\usepackage{bm}
\usepackage{balance}
\usepackage[caption=false,font=normalsize]{subfig}
\usepackage{notoccite}
\usepackage{CJKutf8}
\usepackage{tikz}
\usepackage[compact]{titlesec}
\usepackage{algorithm}
\usepackage{algpseudocode}
\usepackage{stfloats} 
\usepackage{setspace} 

\makeatletter
 \let\NAT@parse\undefined
 \makeatother
\usepackage[pdftex,
        colorlinks=true,
        bookmarks=true,
        citecolor=red,
        linkcolor=blue,
        pdfstartview=Fit,
        breaklinks=true
            ]{hyperref}
\usepackage{cleveref}

\title{\LARGE\bf
PolygMap: A Perceptive Locomotion Framework for Humanoid Robot Stair Climbing
}
\author{Bingquan Li$^{1,2,\dag}$, Ning Wang$^{1,\dag}$, Zhicheng He$^{3}$, Yucong Wu$^{3,4}$, Tianwei Zhang$^{1,5,*}$
\vspace{-1cm}
\thanks{This work was supported by the National Natural Science Foundation of China (Grant No. 62306185); the Guangdong Basic and Applied Basic Research Foundation (Grant No. 2024A1515012065), the Shenzhen Science and Technology Program (Grant No. JSGGKQTD20221101115656029 and KJZD20230923113801004 and ZDCY20250901094531003)}
\thanks{\dag Equal Contribution}
\thanks{$^{1}$The Shenzhen Institute of Artificial Intelligence and Robotics for Society, Shenzhen, China}
\thanks{$^{2}$Guangdong University of Technology, Guangzhou, China}
\thanks{$^{3}$Leju（Shenzhen）Robotics Co. Ltd. Shenzhen, China}
\thanks{$^{4}$Southern University of Science and Technology, Shenzhen, China}
\thanks{$^{5}$The Chinese University of Hong Kong - Shenzhen, Shenzhen, China}
\thanks{* Corresponding Author: {zhangtianwei@cuhk.edu.cn}}
}

\begin{document}
\begin{CJK}{UTF8}{gbsn}
\maketitle
\thispagestyle{empty}
\pagestyle{empty}

\begin{abstract}
Recently, biped robot walking technology has been significantly developed, mainly in the context of a bland walking scheme. To emulate human walking, robots need to step on the positions they see in unknown spaces accurately.
In this paper, we present PolyMap, a perception-based locomotion planning framework for humanoid robots to climb stairs. Our core idea is to build a real-time polygonal staircase plane semantic map, followed by a footstep planar using these polygonal plane segments. These plane segmentation and visual odometry are done by multi-sensor fusion(LiDAR, RGB-D camera and IMUs). The proposed framework is deployed on a NVIDIA Orin, which performs 20-30 Hz whole-body motion planning output. Both indoor and outdoor real-scene experiments indicate that our method is efficient and robust for humanoid robot stair climbing. 
\end{abstract}
\section{Introduction}\label{char1}
Humanoid robots are gradually emerging from laboratories and being deployed in diverse real-world scenarios, including building inspections, emergency responses, and industrial collaborations \cite{caai}. Compared with level-ground locomotion, stairs, a standard yet high-risk man-made structure, present step-height differences, narrow contact surfaces, and fall-prevention constraints, thereby imposing greater demands on the robot's full-chain capability to navigate complex geometries. 
Many existing systems rely on powerful self-balancing and high-bandwidth tracking stabilisers, using rapid whole-body posture regulation to ``brute-force" geometric disturbances during stair climbing \cite{bir}. However, these approaches lack an explicit perception of environmental geometry and foothold regions, often resulting in overly conservative gaits and considerable footstep uncertainty, which makes it difficult to achieve reliable and sustained climbing on long staircases, under varying illumination, or with degraded tread surfaces.

\begin{figure}[htbp]
\centering
\includegraphics[width=0.8\columnwidth]{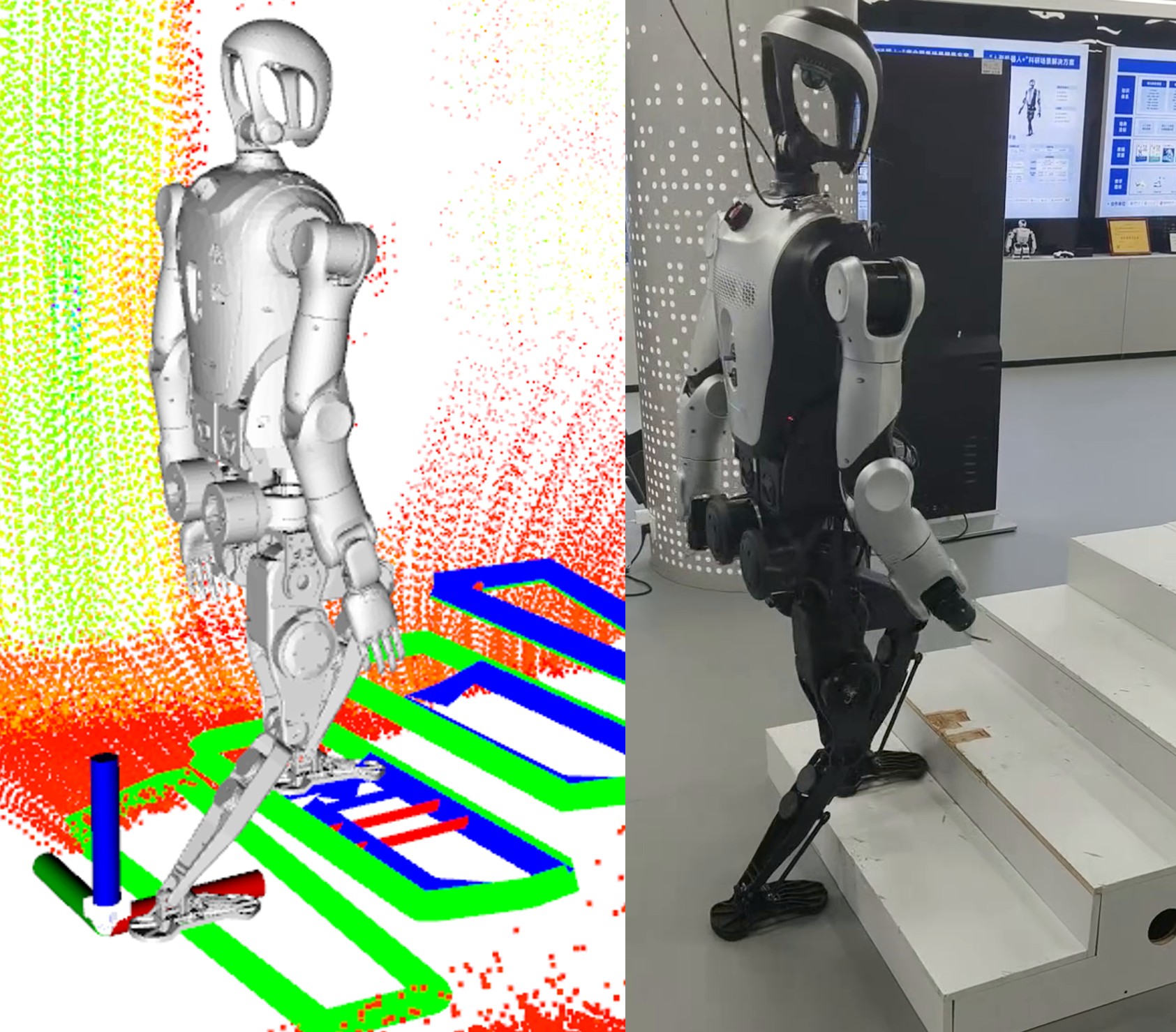}
\caption{Perceptive Locomotion for Humanoid Stair Climbing}
\label{fig:img1}
\end{figure}

Visual perception enables the acquisition of prior information about traversable planes, step edges, and obstacles before foot contact, thereby explicitly constraining footstep planning and mitigating unforeseen risks. 
Perception-based approaches face some challenges in stair environments. First, stair treads are small with sharp edges, and depth measurements are easily affected by low texture, black absorptive materials, and viewpoint changes, resulting in unstable normal estimation and fragmented planes~\cite{ref1}. Second, body vibration during bipedal walking and camera rolling-shutter effects amplify noise and cause sensing data dropouts, making single-frame plane fitting highly sensitive to threshold selection~\cite{8}. Third, odometry–mapping coupling induces local drift that accumulates in multi-frame fusion, degrading the spatio-temporal consistency of step boundaries and traversable polygons. Finally, the perception–planning–control loop must operate in real time, requiring not only GPU-friendly geometric extraction but also map representations that can be efficiently integrated into controllers. 

Consequently, how to robustly generate stair-adaptive traversable region representations under perception noise, state estimation uncertainty and real-time computation constraints, and efficiently couple them with whole-body motion generation, remains a key challenge for achieving reliable stair climbing. 
To this end, we proposed PolygMap: a stair-climbing framework for humanoid robots. See Fig. \ref{fig:img1}, it fuses LiDAR and RGB-D point clouds data (PCD) for staircase mapping and humanoid footstep planning. The codes were open-sourced in \cite{URL-polymap}.
Our main contributions are as follows:
\begin{itemize}
\item A open-sourced perception–planning framework integrating LiDAR with robot forward kinematics to enable real-time online footstep planning for stair climbing.
\item The framework performs precise localisation using LiDAR and fuses visual RGB-D data to extract staircase planes and generate footholds that satisfy stability and safety constraints.
\item  A foot trajectory planner based on safe region selection. The entire framework is validated through both simulation and real-world stair-climbing experiments in indoor and outdoor environments.
\end{itemize}

\section{Related Works}\label{char2}
\subsection{Plane Semantic Mapping}
For legged robot locomotion, research on environment representation has gradually evolved from dense grid maps and elevation maps to bounded planar or polyhedral representations directly for footstep planning. A geometric perception method, the CAPE \cite{3} can simultaneously extract planar and cylindrical features at approximately $300$\,Hz on a single-core CPU, significantly enhancing the geometric constraints of visual odometry. In~\cite{5}, a hierarchical clustering and improved flood-filling algorithm were proposed for multi-plane segmentation at around $35$\,Hz. To satisfy the real-time requirements, in~\cite{81}, A GPU-based pipeline for planar region extraction and elevation mapping was proposed, achieving $150$–$200$\,Hz output rates for local planar regions and providing height layers with uncertainty estimates for navigation and gait planning.
Several studies have demonstrated real-time performance exceeding $100$\, Hz~\cite{6} and improved the accuracy and robustness of footstep solutions by coupling with kinematic–inertial estimation~\cite{stephane19}. 

Overall, research on planar semantic mapping is moving from local geometric descriptions toward structured and spatio-temporally consistent models of feasible foothold regions, enabling tight coupling with footstep planning and whole-body control.

\subsection{Perspective Locomotion}
In the early stages, people typically treat humanoid robot footstep planning and walking pattern generation as separate planning questions. 
For example, Roychoudhury et al. proposed searching for fast footsteps using 3D polygonal mapping \cite{10}.
However, such an RRT-based method met the real-time efficiency challenge in on-board systems. Then, during DRC-2015, many teams adopted the multi-sensor fusion strategy for on-board perspective footstep planning and motion control.  For instance, in \cite{fallon2014}, the MIT team fuses LIDAR, inertial and kinematics for the Atlas robot's rough terrain walking task.

In recent years, Elevation map~\cite{18} has been widely applied in quadruped locomotion. Its probabilistic terrain modelling explicitly represents localisation errors and confidence bounds, providing optimism with more reliable constraints.
Griffin et al. \cite{griffin2024} introduced an efficient terrain mapping algorithm that integrates planar region measurements with kinematic–inertial state estimation. 
Building upon these advances, perceptive locomotion further incorporates environmental geometry directly into the online optimisation of contact sequences and whole-body motions. In~\cite{hutter-TRO23}, the nonlinear MPC first performs traversability classification, plane segmentation, and signed distance field preprocessing on elevation maps, transforming terrain constraints into convex inequalities. It has been validated in scenarios such as gap crossing, stepping-stone traversal, and slope walking. 
In~\cite{SR25}, a learning-based method is proposed that utilises attention-based map encoding, allowing end-to-end control policies to focus on future traversable regions automatically. This improves robustness to uncertainty and enables traversal of sparse footholds, with real-world validation on both quadruped and humanoid platforms. 

\section{Methods}
\label{char3}
\begin{figure*}[htbp]
\centering
\includegraphics[width=0.8\textwidth]{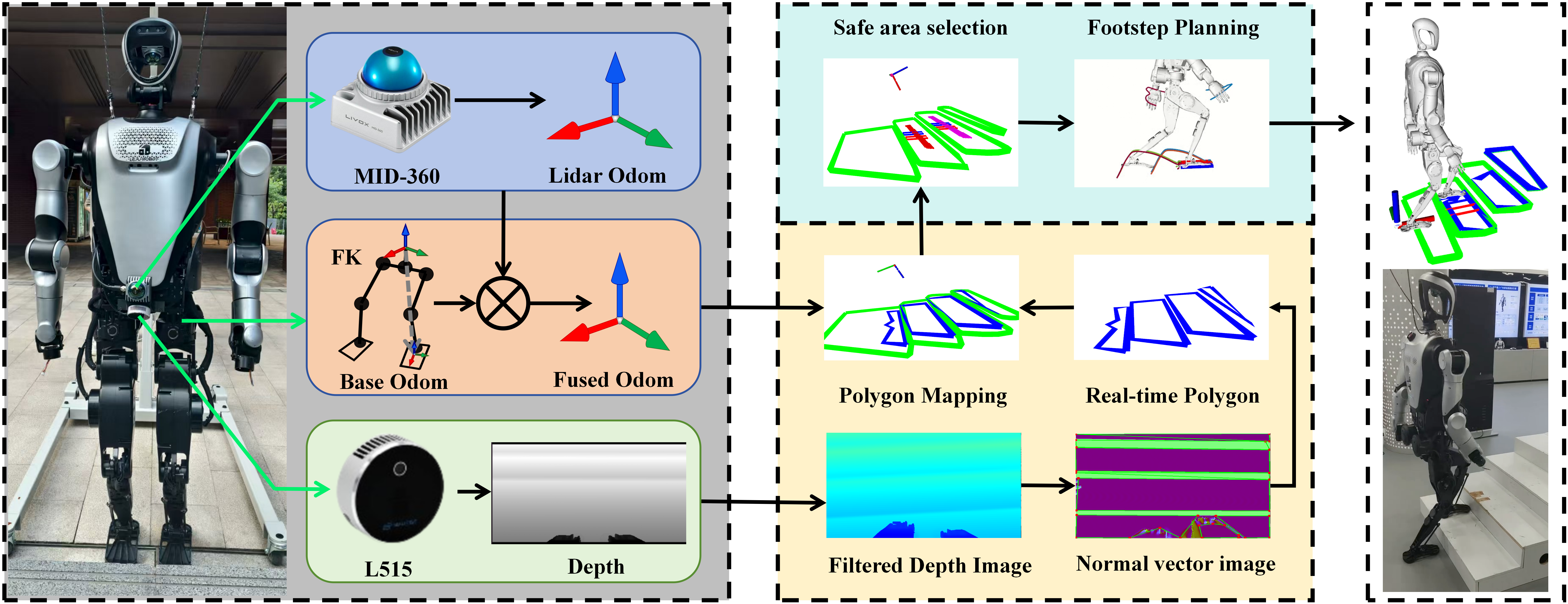}
\caption{The system integrates joint recorders, depth sensing and LIO estimator. Robot pose is obtained via fusing forward kinematics and the LIO, while depth images provide a polygonal map for foothold generation. Based on these footholds, footstep and whole body motion planning enable stable stair climbing.}
\label{fig:overview_of_system}
\vspace{-0.6cm}
\end{figure*}

The proposed perceptive locomotion framework is illustrated in Fig.~\ref{fig:overview_of_system}. As described on the left side of the picture, the sensor configuration enables the robot to perform navigation and mapping in complex, unknown stair environments. The downward-facing Realsense L515 leverages LiDAR technology to capture depth data even under low-light conditions, though its performance degrades on highly absorptive surfaces. For odometry and pose estimation, a Livox Mid360 LiDAR and an IMU are employed, with Point-LIO providing accurate odometry estimation.

\begin{figure}[tbhp]
\centering
\includegraphics[width=0.75\columnwidth]{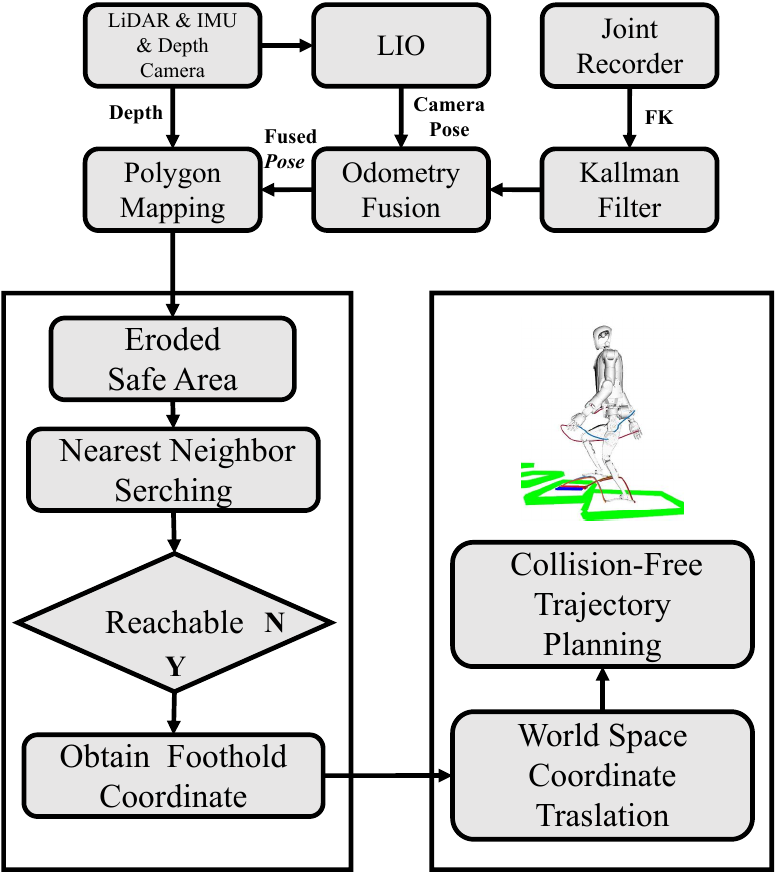}
\caption{Polygmap-based footstep motion planning logic}
\label{fig:overview_of_processing}
\end{figure}

This framework leverages real-time semantic planes for staircase representation and motion planning, aiming to achieve accurate pose estimation and robust climbing motion control on uneven surfaces. Fig.~\ref{fig:overview_of_processing} illustrates our processing pipeline. The system inputs include depth images from an RGB-D camera, initial pose estimates from LiDAR-Inertial Odometry (LIO), and kinematic pose estimates obtained via forward kinematics. 
Xinyi et al. proposed to filter moving object points in LIO front end \cite{xinyi}, for biped walking robots, their own vibrations result in more dynamic points for the LIO. 
We first adopt an anisotropic diffusion filtering from \cite{binteng} for depth image processing.
Then, the surface normal maps are computed using Sobel operators, contours are extracted with the Canny algorithm, and polygons are fitted to the environment using RANSAC. In parallel, robot joint data are combined with forward kinematics to generate a smooth body state estimate, which is then fused with the LIO pose to construct a polygonal semantic map. Using this simplified polygon map, the safe regions are eroded to extract feasible foothold candidates and assess their reachability. The footholds are then generated with collision-free trajectories. 

\subsection{Multi-Sensor Fusion based State Estimation}
This section details the smooth and accurate robot pose estimation methodology integrating proprioceptive sensing and LiDAR-Inertial Odometry (LIO). We adopt a linear Kalman filter to integrate multi-modal data from the IMU, joint encoders, and foot contact states, thereby enabling a unified estimation of the robot's base position, velocity, orientation, and foot positions. Additionally, a loosely coupled scheme is employed to integrate pose estimates from LIO into the fusion framework, enhancing the overall system's robustness and accuracy.

\textbf{Forward Kinematics:}
Assuming the robot state variables are:
\begin{equation}
\mathbf{x} = \begin{bmatrix}
{^W}\mathbf{p}_{\text{base}}^\top & {^W}\mathbf{v}_{\text{base}}^\top & {^W}\mathbf{p}_{\text{c}_1}^\top & \cdots & {^W}\mathbf{p}_{\text{c}_8}^\top
\end{bmatrix}^\top \in \mathbb{R}^{30}
\end{equation}
where ${}^{W}\!\mathbf{p}_{\mathrm{base}},\;{}^{W}\!\mathbf{v}_{\mathrm{base}}$ represent the position and velocity of the robot's CoM in the world coordinate frame $W$;
${}^{W}\!\mathbf{p}_{c_i}$ represents the contact points of the feet (four per foot, eight in total), i.e., $i=1,\dots,8$. The state-space equation is:

\begin{equation*}
\mathbf{x}_{k+1} = \mathbf{A}\mathbf{x}_k + \mathbf{B}\mathbf{u}_k
\end{equation*}
\begin{equation}
\mathbf{A} = \begin{bmatrix}
\mathbf{I}_3 & \Delta t \cdot \mathbf{I}_3 & \mathbf{0} \\
\mathbf{0} & \mathbf{I}_3 & \mathbf{0} \\
\mathbf{0} & \mathbf{0} & \mathbf{I}_{24}
\end{bmatrix}
\quad
\mathbf{B} =
\begin{bmatrix}
0.5 \Delta t^2 \cdot \mathbf{I} \\
\Delta t \cdot \mathbf{I} \\
\mathbf{0}_{24} \\
\end{bmatrix}
\end{equation}
where $\mathbf{I} = \mathbf{I}_{3 \times 3}$, and the blank entries represent $\mathbf{0}_{3 \times 3}$. The control input u is the linear acceleration of the robot's CoM in the world frame, i.e., $\mathbf{u} = {^W}\mathbf{a}_{\text{base}}$. The observation vector is denoted as:
\begin{equation}
\mathbf{y}=\begin{bmatrix}{^B}\mathbf{p}_c\\{^B}\mathbf{v}_c\\\mathbf{h}_c\end{bmatrix}\in\mathbb{R}^{56}.
\end{equation}
where ${}^{B}\!\mathbf{p}_{c},\;{}^{B}\!\mathbf{v}_{c}$ represent the position and velocity of the foot contact points relative to the robot's CoM, and $\mathbf{h}_c$ are the height of the contact points, with
\begin{equation}
{^B}\mathbf{p}_c=\begin{bmatrix}{^B}\mathbf{p}_{c,1}\\\vdots\\{^B}\mathbf{p}_{c,8}\end{bmatrix},\quad
{^B}\mathbf{v}_c=\begin{bmatrix}{^B}\mathbf{v}_{c,1}\\\vdots\\{^B}\mathbf{v}_{c,8}\end{bmatrix},\quad
\mathbf{h}_c=\begin{bmatrix}h_{c,1}\\\vdots\\h_{c,8}\end{bmatrix}
\qquad
\end{equation}

The robot torso state (position and velocity) is then estimated in real-time with a linear Kalman filter that fuses LiDAR odometry, proprioceptive measurements, and IMU orientation, yielding smooth and low-drift odometry for the subsequent perception–planning–control loop.

\textbf{Odometry Fusion:}
 Given the world coordinate frame $W$, the robot body frame $\text{B}$, the LIO estimation ${}^{W}\mathbf T_{L}\in SE(3)$, and the robot body odometer ${^{W}}\mathbf{T}_{\text{base}}$. Since the LiDAR is rigidly connected to the robot, the extrinsic parameters from the robot's CoM to the LiDAR ${^B}\mathbf{T}_{\text{lidar}}$ are pre-known. Then, to transform LIO and the camera's odometry to the robot's CoM:
\begin{equation}
{^W}\mathbf T_{\mathrm{base}}^{(L)} = {^W}\mathbf T_{\text{lidar}} \left(^{\mathrm{B}} \mathbf T_{\text{lidar}}\right)^{-1}.
\end{equation}
The LIO odometry is obtained through a laser SLAM algorithm (Point-LIO \cite{he2023point}). 
The complementary filtering for the position part is expressed as:
\begin{equation}
{^W\mathbf p_{\text{base}}^{\text{(fused)}}} = \alpha \, {^W\mathbf p_{\text{base}}} + (1-\alpha)\, {^W \mathbf p_{\text{base}}^{(L)}}
\end{equation}
where $\alpha \in [0,1]$ is the complementary coefficient. Attitude fusion can be achieved through the exponential map form:
\begin{equation}
\Delta R = \log\!\Big( \big({^WR_{\text{base}}}_k\big)^\top {^{W}R_{\text{base}}^{(L)}}_k \Big)
\end{equation}
\begin{equation}
^W R_{\text{base}}^{\text{fused}} = {^{W}R_{\text{base}}}_k \exp\!\big( (1-\alpha)\,\Delta R \big),
\end{equation}
The relationship between the complementary coefficient $\alpha$ and the time constant $\tau$ is:
\begin{equation}
\alpha = \frac{\tau}{\tau+\Delta t}, \qquad 1-\alpha = \frac{\Delta t}{\tau+\Delta t},
\end{equation}
Finally, a precise and smooth robot odometry estimation is obtained:
\begin{equation}
{^W}\mathbf T_{\mathrm{base}}^{\text{(fused)}} = \left({^W\mathbf p_{\text{base}}^{\text{(fused)}}},{^{W}R_{\text{base}}^{\text{fused}}} \right)
\end{equation}
\subsection{Foothold Generation}
During polygonal map construction, the fused odometry and RGB-D depth image are applied to extract a polygonal representation of the staircases. Each polygon undergoes convex hull computation and rasterisation. Robot foot unreachable regions are removed using range filtering, and the isolated or unstable edge points are suppressed through layered erosion operations, resulting in a set of stable candidate foothold regions. Based on these candidates, we select the optimal foothold according to distance and height criteria, and then publish them to the downstream foothold planner for trajectory generation and execution.

\textbf{Polygon Map:}
To construct the Polygon Map, we first calculate the normals $\vec{n}(u,v)$ of each pixel from the depth image and camera intrinsics.

\begin{equation}
P(u,v)=
\begin{bmatrix}
\frac{(u-c_x)D(u,v)}{f_x},\ 
\frac{(v-c_y)D(u,v)}{f_y},\ 
D(u,v)
\end{bmatrix}^T .
\end{equation}
Within the pixel neighborhood, differential vectors are constructed:
$\vec{a}=P(u+1,v)-P(u,v)$ and
$\vec{b}=P(u,v+1)-P(u,v)$.
Their cross product is normalized to obtain the local normal vector:
\begin{equation}
\vec{n}(u,v)=
\frac{\vec{a}\times\vec{b}}
{\|\vec{a}\times\vec{b}\|}.
\end{equation}
This directly yields the normal vector map without the need to explicitly generate a dense PCD.
Meanwhile, to reduce depth noise and preserve edge features, an anisotropic diffusion filter is adopted in iterative form:
\begin{equation}
I^{t+1}(x,y)=I^t(x,y)+\lambda\sum_{i\in \mathcal{N}(x,y)} c_i(x,y)\,\nabla I_i(x,y),
\end{equation}
where the thermal conductivity coefficient $c_i(x,y)$ is adaptively adjusted based on the gradient magnitude to smooth background regions while protecting edges. 

Subsequently, within the extracted polygonal areas, we apply the RANSAC algorithm for plane fitting, iteratively obtain the optimal plane parameters. This process implements the key steps from raw depth data to robust plane extraction, providing the foundation for subsequent map construction and gait planning.


\textbf{Optimal Foothold:}
To extract candidate footholds for the robot from the Polygon Map, we assume the received polygon set is $\mathcal{M} = \{M_i\}$, where each polygon $M_i$ consists of a vertex set $\{p_{i,j}\}_{j=1}^{N_i}$, with each vertex containing spatial coordinates $(x_{i,j},y_{i,j},z_{i,j})$. To ensure the stability of the 2D convex hull calculation, the height of each vertex is set to zero:
\begin{equation}
P_i = \{(x_{i,j},y_{i,j},0)\}, \quad j=1,\dots,N_i
\end{equation}

Subsequently, the 2D convex hull $H_i$ of $P_i$ is computed:
\begin{equation}
H_i = \operatorname{ConvexHull}(P_i), \quad H_i \subset \mathbb{R}^2
\end{equation}
which describes the outer bounding contour of the polygon in the XY plane (horizontal). To generate a dense PCD inside the convex hull, calculate its bounding box:
\begin{equation}
[x_{\min}, x_{\max}] \times [y_{\min}, y_{\max}] = \text{BoundingBox}(H_i)
\end{equation}

A uniform grid of points $(x,y)$ is generated within this bounding box with resolution $g_{\text{res}}$. Only points located inside the convex hull are kept:
\begin{equation}
(x,y) \in H_i
\end{equation}
while their height is assigned the average value of the original polygon vertices:
\begin{equation}
z = \frac{1}{N_i} \sum_{j=1}^{N_i} z_{i,j}
\end{equation}

This yields the set of dense PCD for all polygons:
\begin{equation}
P_{\text{total}} = \bigcup_i \{(x,y,z)\}
\end{equation}
To ensure the validity of footholds near the robot's base coordinate frame, coordinate transformation and range filtering are applied to the PCD. Denote the robot base coordinate as \texttt{base\_link}. After transforming the PCD from the world frame $W$ to the robot body frame $B$, keep the points within the XY plane distance:
\begin{equation}
|x(p)| \le g_{\text{range}}, \quad |y(p)| \le g_{\text{range}}, \quad p \in P_{\text{total}}
\end{equation}
The filtered PCD is denoted as $P_{\text{filtered}}$.
To extract the salient height features of the planar PCD, we extract the maximum height point of each grid (side length $g_{\text{res}}$):
\begin{equation}
P_{\text{max}} = \bigcup_{\text{grid } G} \arg\max_{p \in G} z(p)
\end{equation}
thereby obtaining the set of highest points for each XY grid, facilitating subsequent foothold analysis and environment modelling. Consider the limitation of the robot's sole height; the minimum sole height is first obtained from the transformations of both toes and heels with respect to the body's CoM:
\begin{equation}
z_{\text{foot}} = \min\{ z_{\text{lltoe}}, z_{\text{rrtoe}}, z_{\text{llheel}}, z_{\text{rrheel}} \}
\end{equation}
Subsequently, points higher than the foot threshold $g_z$ are removed:
\begin{equation}
P_{\text{max\_filtered}} = \{ p \in P_{\text{max}} \mid z(p) \le z_{\text{foot}} + g_z \}
\end{equation}
To remove isolated points and edge noise, $P_{\text{max\_filtered}}$ is divided into height layers, with each layer height being $h_{\text{layer}}$:
\begin{equation}
L_k = \{ p \in P_{\text{max\_filtered}} \mid k = \lfloor z(p)/h_{\text{layer}} \rfloor \}
\end{equation}
and $N_{\text{erosion}}$ erosion operations are performed on each layer to filter out sharp edge points. The PCD from all layers after erosion are merged to form the final structured PCD:
\begin{equation}
P_{\text{eroded}} = \bigcup_k L_k^{\text{eroded}}
\end{equation}
Within $P_{\text{eroded}}$, the point closest to the robot's base coordinate that also satisfies the height condition for stepping is selected as the primary foothold candidate:
\begin{equation}
\begin{aligned}
& p \in P_{\text{eroded}},\, z(p) > z_{\text{foot}} + \Delta_{\text{foot}}\\
& p^* = \arg\min \| (x(p)-x_{\text{base}}, y(p)-y_{\text{base}}) \|^2
\end{aligned}
\end{equation}
To reflect the height distribution of nearby obstacles, the next-step foothold solution point $p^{**}$, higher than $p^*$, could also be selected. The relative direction of the candidate point $p^*$ with respect to the orientation $\psi_{\text{base}}$ of the robot's base coordinate is:
\begin{equation}
\theta_{\text{rel}} = \arctan2(y(p^*), x(p^*)) - \psi_{\text{base}}, \quad \theta_{\text{rel}} \in [-\pi, \pi]
\end{equation}
Furthermore, the PCD layer near the sole height is:
\begin{equation}
z_{\text{foot}} - \Delta_{\text{foot}} \le z(p) \le z_{\text{foot}} + \Delta_{\text{foot}}
\end{equation}
We apply separate erosion processing to this plane to improve the climb locomotion safety. 
Finally, the optimal foothold points are generated around the candidate points $p^*$ and $p^{**}$. This method converts sparse 2D polygon markers into a structured, dense PCD, providing both spatial position and orientation for robot footstep planning.

\subsection{Footstep Planning}
In this subsection, we propose a bipedal gait trajectory generation method based on real-time robot pose estimation. The system input is the estimated pose of the robot, denoted as:
\begin{equation}
    ^B \mathbf{t} = [x, y, z, \phi]^\mathrm{T}
\end{equation}
where \((x, y, z)\) represents the spatial position of the torso, and \(\phi\) is the yaw orientation. This method aims to generate smooth and executable footstep trajectories:
\begin{equation}
   ^B \mathbf{T}_{\mathrm{foot}} = \{ (\mathbf{p}_i^f, \mathbf{p}_i^t, t_i) \}_{i=1}^N
\end{equation}
where \(\mathbf{p}_i^f \in \mathbb{R}^3\) represents the foot position of the $i$-th step, $\mathbf{p}_i^t \in \mathbb{R}^4$ represents the torso pose, and $t_i$ is the corresponding time.
The foot pose is generated from the torso pose $\mathbf{p}_t = [x_t, y_t, z_t, \phi_t]$ and the left/right foot offset $\mathbf{b}_f = [0, \pm y_b, -z_t]$. Using the rotation matrix $\mathbf{R}_z(\phi_t)$, the offset is rotated to the torso direction to obtain the foot position:
\begin{equation}
   \mathbf{p}_f = \mathbf{p}_t + \mathbf{R}_z(\phi_t) \cdot \mathbf{b}_f;\
\mathbf{R}_z(\phi_t) = 
\begin{bmatrix}
\cos\phi_t & -\sin\phi_t & 0\\
\sin\phi_t & \cos\phi_t & 0\\
0 & 0 & 1
\end{bmatrix}
\end{equation}

To ensure the smoothness of the gait, the foot trajectory along the Z-direction (vertical) during the lift-off and landing phases is interpolated using a sinusoidal curve:
\begin{equation}
z_f(t) = 
\begin{cases}
z_{\max} \sin \left( \dfrac{\pi t}{2 t_{\mathrm{lift}}} \right) & 0 \le t \le t_{\mathrm{lift}} \\
z_{\max} \cos \left( \dfrac{\pi (t - t_{\mathrm{lift}})}{2 t_{\mathrm{land}}} \right) + z_0 & t_{\mathrm{lift}} < t \le T
\end{cases}
\end{equation}
where $T = t_{\mathrm{lift}} + t_{\mathrm{land}}$ represents the total swing phase duration, $z_{\max}$ is the maximum step height, $t_{\mathrm{lift}}$ and $t_{\mathrm{land}}$ are the time durations for the lift-off and landing phases, respectively, and $z_0$ is the reference landing height.
\begin{equation}
 x_f(t) = x_{\mathrm{start}} + (x_{\mathrm{end}} - x_{\mathrm{start}}) \frac{t}{t_{\mathrm{step}}}   
\end{equation}

To avoid foot overlap, a rotating rectangle model is used to represent the occupied space of the foot:
\begin{equation}
\begin{aligned}
&\mathcal{R}_f = \mathrm{RotatingRectangle} (\mathrm{center}=(x_f, y_f),
\\
&\mathrm{width}=w, \mathrm{height}=h, \theta=\phi_t)
\end{aligned}
\end{equation}
and the step sequence is adjusted by checking whether the rectangle of the current step intersects with the previous foot. If both feet intersect, a trajectory conflict is indicated, and the input torso path needs to be adjusted.

To determine the swing leg sequence, a motion-consistent selection rule is applied.
Let the torso displacement expressed in the previous local frame be
\begin{equation}
\Delta \mathbf{p}_t =
\mathbf{R}_z(\phi_{t-1})^\top
\left(
\mathbf{p}_t - \mathbf{p}_{t-1}
\right)
\end{equation}
The swing leg at step $i$ is selected as
\begin{equation}
\mathrm{foot}_i =
\begin{cases}
\text{left}, & \phi_t > 0 \ \text{or} \ \Delta y_t > 0 \\
\text{right}, & \text{otherwise}
\end{cases}
\end{equation}
where $\Delta y_t$ denotes the lateral component of $\Delta \mathbf{p}_t$.

The gait trajectory is discretised at fixed time intervals $\Delta t$:
\begin{equation}
t_i = i \Delta t, \quad i = 1, \dots, N    
\end{equation}
The entire generation process achieves torso-foot coordination, smooth foot lifting, and collision-safe bipedal gait planning.

\section{Experiments Results and Evaluations}
To validate the effectiveness of the proposed perceptive staircase locomotion method,  The robot employed in the experiments is a KUAVO humanoids (see Fig.~\ref{fig:img1}). It is 166\,cm high, 55\,kg weight and 28 degrees of freedom. The foot size is 26\,$\times$\,9.6\,cm. To perform the proposed perception method, we integrated a 3D LiDAR, an RGB-D camera and an IMU. All computations were carried out on an NVIDIA Orin NX, enabling whole-body motion planning to be performed in real time at a frequency of 20–30\,Hz.

Three groups of experiments were designed. First, a sensor fusion experiment was conducted to verify the localization accuracy of multi-sensor fusion. Then, a Gazebo simulation experiment was carried out to evaluate the stability and real-time performance of the perception--planning--control loop under ideal conditions. Finally, real-world experiments were performed in both indoor and outdoor environments to assess the reliability and overall accuracy of the proposed method.
The evaluation metrics include: the continuity of stair detection, the number of steps that can be climbed consecutively, and the horizontal and vertical errors between the planned and executed footholds. 
\subsection{Sensor Fusion Experiment Analysis}
\begin{figure}[thbp]
\centering
\includegraphics[width=\columnwidth]{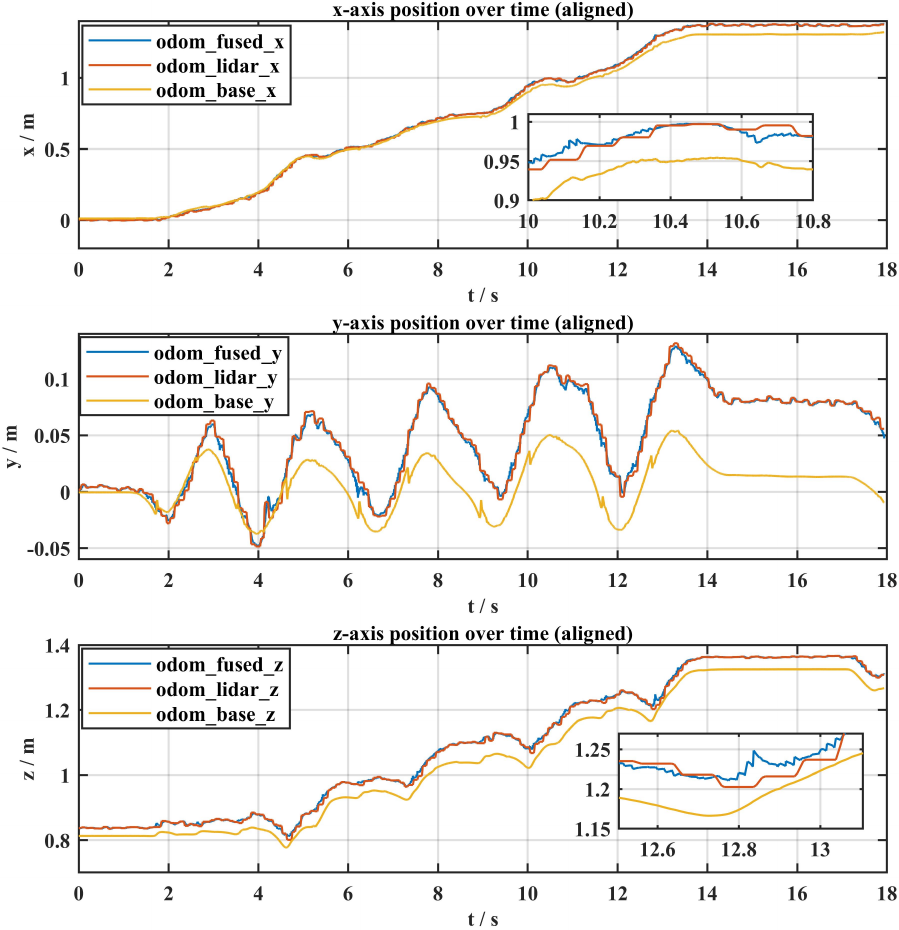}
\caption{Comparative Analysis of Position Estimation: Ontology-Based, LIO, and the proposed Fusion Method}
\label{fig:odom3}
\vspace{-0.3cm}
\end{figure}

Fig.~\ref{fig:odom3} illustrates the trajectories of three odometry estimations (odom\_base, odom\_lidar, and odom\_fused) along the $x$, $y$, and $z$ axes during humanoid robot walking. The odom\_base estimation, which relies solely on proprioceptive sensors, exhibits significant cumulative drift over time, particularly along the $x$- and $z$-axes, leading to large localization errors during extended operation. In contrast, the LiDAR-based odometry (odom\_lidar) provides higher absolute positioning accuracy but suffers from pronounced high-frequency oscillations, especially in the $y$-axis direction, making it unsuitable for direct use in the control module.  

The fused result (odom\_fused) combines the advantages of both estimations: it retains the global accuracy of LIO while incorporating proprioceptive observations to constrain and smooth the state estimation, resulting in much more stable trajectories along all three axes. The zoomed-in insets further demonstrate that the fused trajectory effectively suppresses the high-frequency noise of odom\_lidar and significantly reduces cumulative drift. These results indicate that our multi-sensor fusion method generates accurate and smooth state estimates, providing reliable pose information for subsequent stair detection and foothold planning.  

\subsection{Gazebo Simulation Results}
Table~\ref{tab:sim_results} presents five sets of simulation data obtained in the Gazebo simulation, where three sets correspond to a 10-level staircase and two sets to a 4-level staircase. The table reports the total climbing time, number of steps, plane detection frequency, and maximum footstep error. The results show that the most extended duration from the start of climbing until the robot reaches a stable stance is 35\,s, with an average time of approximately 3.4--3.7\,s per step. This indicates that the system achieves relatively high motion efficiency while maintaining stability. During the stair-climbing process, the maximum footstep error from detection to execution is 12.4\,mm, which is significantly smaller than the available tread width. This satisfies the required safety margin between the foot size and the stair tread, ensuring stable climbing performance.  

\begin{table}[htbp]
\centering
\caption{Gazebo Simulation Results}
\label{tab:sim_results}
\begin{tabular}{c c c c c}
\hline
\textbf{Trial} & \textbf{$T$ (s)} & \textbf{Steps} & \textbf{$FPS$(Hz)} & \textbf{$e_{m}(mm)$} \\
\hline
1 & $34$ & $10$ & $21$ & $11.2$ \\
2 & $35$ & $10$ & $20$ & $10.3$ \\
3 & $34$ & $10$ & $25$ & $12.0$ \\
4 & $14$ & $4$ & $29$ & $12.4$ \\
5 & $15$ & $4$ & $28$ & $11.8$ \\
\hline
\textbf{max} & 35 & 10 & 29 & 12.4 \\
\hline
\end{tabular}
\vspace{-0.3cm}
\end{table}

\begin{figure}[tbhp]
    \centering
    \includegraphics[width=0.8\columnwidth]{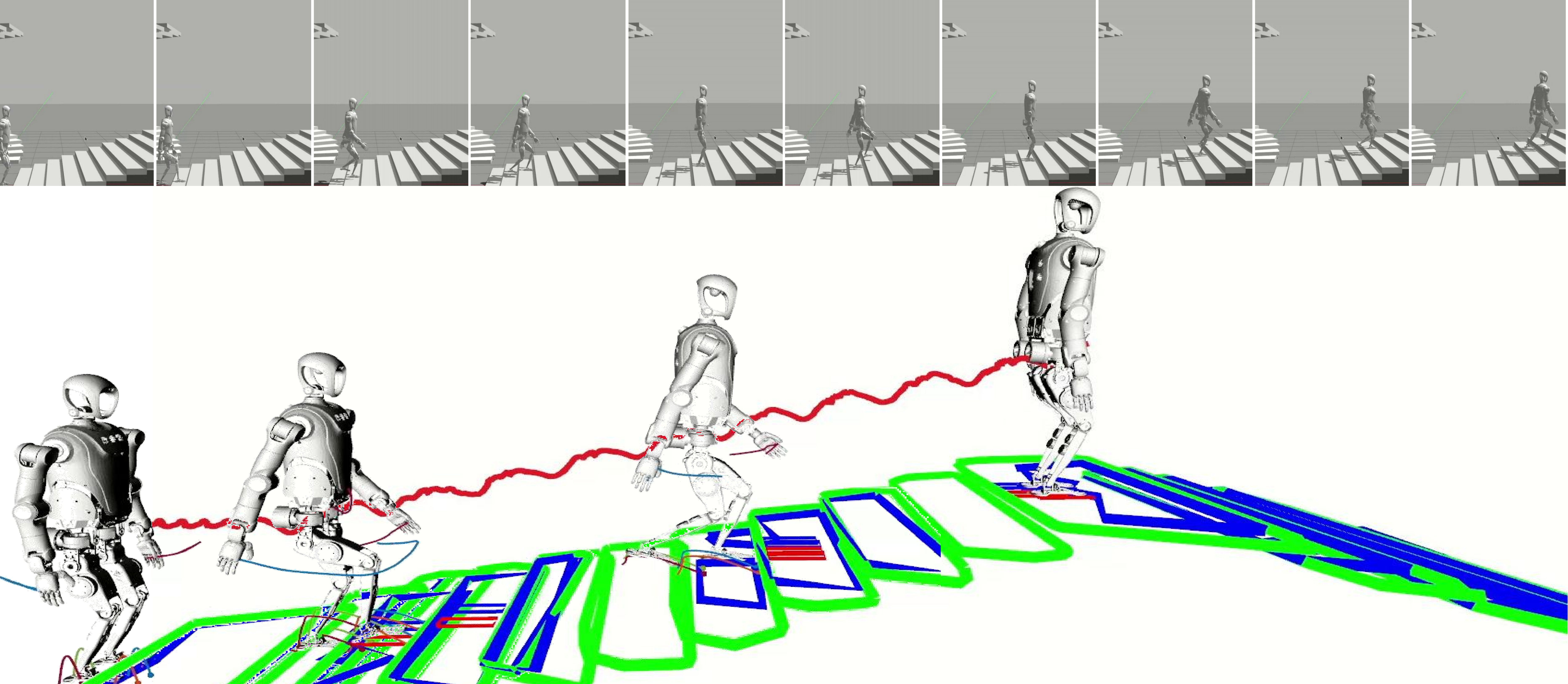}
    \caption{Simulation experiment of the proposed system. The figure shows the perception--planning--control pipeline and stair climbing locomotion sequence.}
    \label{fig:fangzhen_1}
    \vspace{-0.5cm}
\end{figure}

Fig.~\ref{fig:fangzhen_1} illustrates the snapshots of Trial 1. In which the global trajectories of the robot (CoM, hand joints, foot boundaries) are shown. The robot starts from the ground plane and climbs step by step until reaching the upper platform. The entire process is smooth, and the foot placements are accurate, demonstrating the effectiveness and stability of the proposed perception--planning--control loop in simulation. These results provide valuable reference for parameter tuning and controller configuration in subsequent real-world experiments.  

\begin{figure}[htbp]
    \centering
    \includegraphics[width=0.5\textwidth]{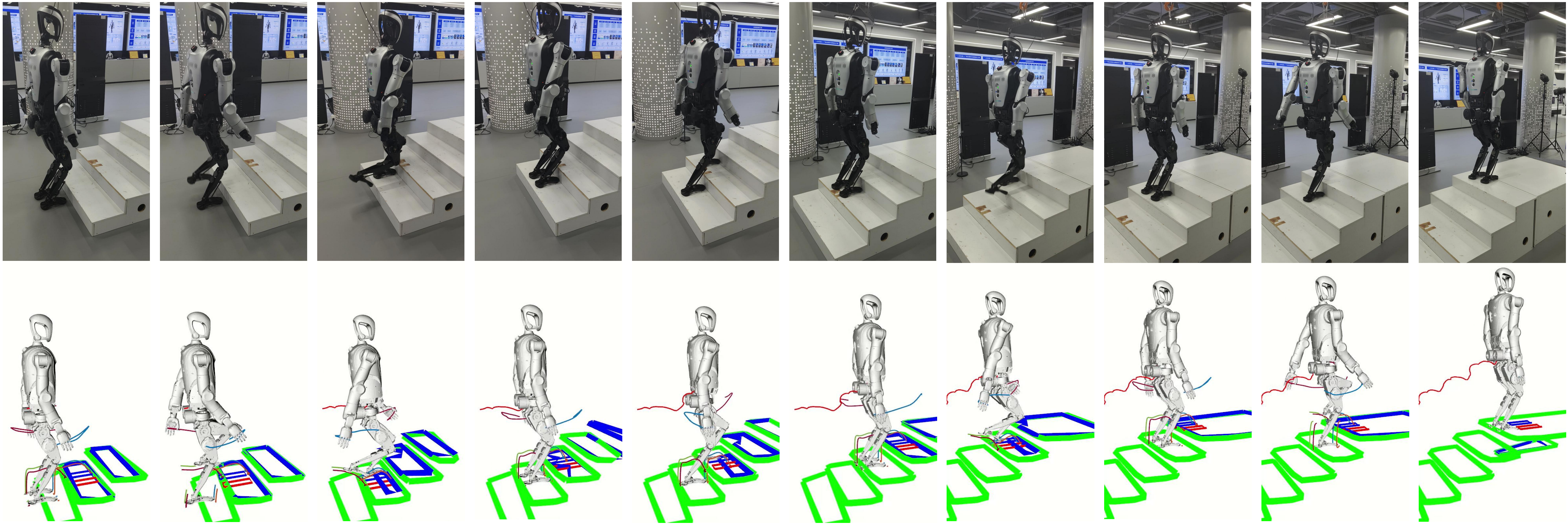}
    \caption{Real-world stair-climbing experiment. The figure presents the experimental setup and key motion snapshots during the perception--planning--control process.}
    \label{fig:shiwu_1}
\end{figure}

\subsection{Real World Stair Climbing Evaluations}
Table~\ref{tab:shiwu_results} reports the data of five times real-world stair-climbing experiments conducted both indoors and outdoors. Two gait strategies were designed: \textbf{DS (Double Step) } indicates that both feet step onto the same stair before proceeding to the next one, whereas \textbf{SS (Single Step)} represents a continuous stepping gait in which each foot alternately steps onto successive stairs, resulting in a faster forward pace. The table summarises the total climbing time, number of steps, plane detection frequency, and maximum footstep error for each trial.  
\begin{figure*}[hbp]
    \centering
    \includegraphics[width=0.8\textwidth]{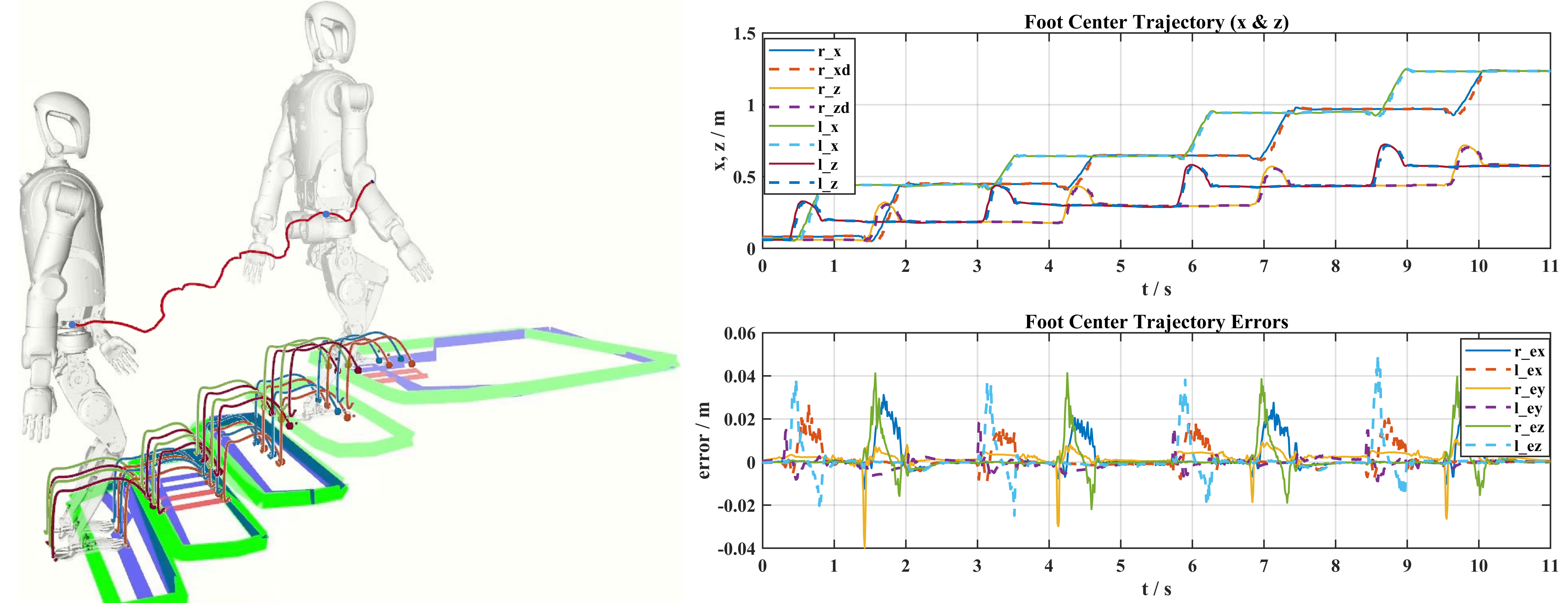}
    \caption{Experimental results of real-world stair climbing. The left plot shows the 3D trajectory of the robot base and footsteps, while the right plot illustrates the foot trajectory tracking performance along the $x$- and $z$-axes, including tracking error curves.}
    \label{fig:shiwu_2}
\end{figure*}

\begin{table}[htbp]
\centering
\caption{Real World Stair Climbing Experiment Results}
\label{tab:shiwu_results}
\begin{tabular}{c c c c c c}
\hline
\textbf{Scene} & \textbf{Gait} & \textbf{$T$ (s)} & \textbf{Steps} & \textbf{$f$(Hz)} & \textbf{$e_{m}(mm)$} \\
\hline
Indoor  & DS & $9.6$ & $4$ & $21$ & $12.1$ \\
Indoor  & DS & $10.2$ & $4$ & $20$ & $11.4$ \\
Indoor & SS & $7.7$ & $4$ & $23$ & $24.7$ \\
Outdoor & SS & $11.2$ & $6$ & $20$ & $33.4$ \\
Outdoor & SS & $9.8$ & $5$ & $21$ & $22.2$ \\
\hline
\textbf{max} & -- & 11.2 & 6 & 23 & 33.4\\
\hline
\end{tabular}
\vspace{-0.5cm}
\end{table}

The results show that the DS gait requires a total time of approximately 9.6--10.2\,s. Each step is planned after the robot reaches a stable stance, yielding a smooth and steady gait with smaller footstep errors. In contrast, the SS gait, though faster, performs online planning while walking, which may prevent the robot from consistently obtaining stable base observations. This leads to generally larger footstep errors, with the maximum reaching 33.4\,mm, approaching the safety margin between the foot size and the stair tread. In outdoor scenarios, despite ground unevenness and illumination variations, the plane detection frequency remains within 20--23\,Hz, which is sufficient to meet real-time planning requirements. These findings demonstrate that the proposed method maintains good robustness in real-world environments.  

Fig.~\ref{fig:shiwu_1} illustrates a complete process of one real stair climbing experiment. This indoor trial was conducted in a staircase model with standard dimensions (step height of 13\,cm and tread width of 28\,cm, which is quite challenging for our robot with a 26\,cm length foot).
Using the DS gait strategy, the robot successfully climbed a total of four steps. At the beginning of the experiment, the robot stood at the bottom of the staircase. After completing environment perception and stair detection through multi-sensor fusion, foothold candidates were gradually generated, and whole-body trajectories were planned. The robot then executed the first climbing step according to the scheduled foot target pose. After reaching a stable double-support stance, foothold updates and trajectory re-planning were performed to complete the four-step climbing task sequentially.  

The upper row of Fig.~\ref{fig:shiwu_1} shows snapshots of the real experimental scene, where the robot steadily performs stepping, support, CoM transfer, and double-support phases in sequence. The lower row presents the corresponding RViz visualisation, where the \textbf{green} rectangles indicate the detected stair planes, the \textbf{blue} rectangles represent the feasible foothold planning regions, and the \textbf{red} and \textbf{purple} small rectangles denote the planned footstep targets for the next level stair. The trajectory curves visualise the motion of the robot's torso, arms, and feet. The entire climbing process is smooth and stable, confirming the effectiveness and reliability of the proposed perception--planning--control loop under the DS gait strategy.

\begin{figure}[tbhp]
    \centering
    \includegraphics[width=0.95\columnwidth]{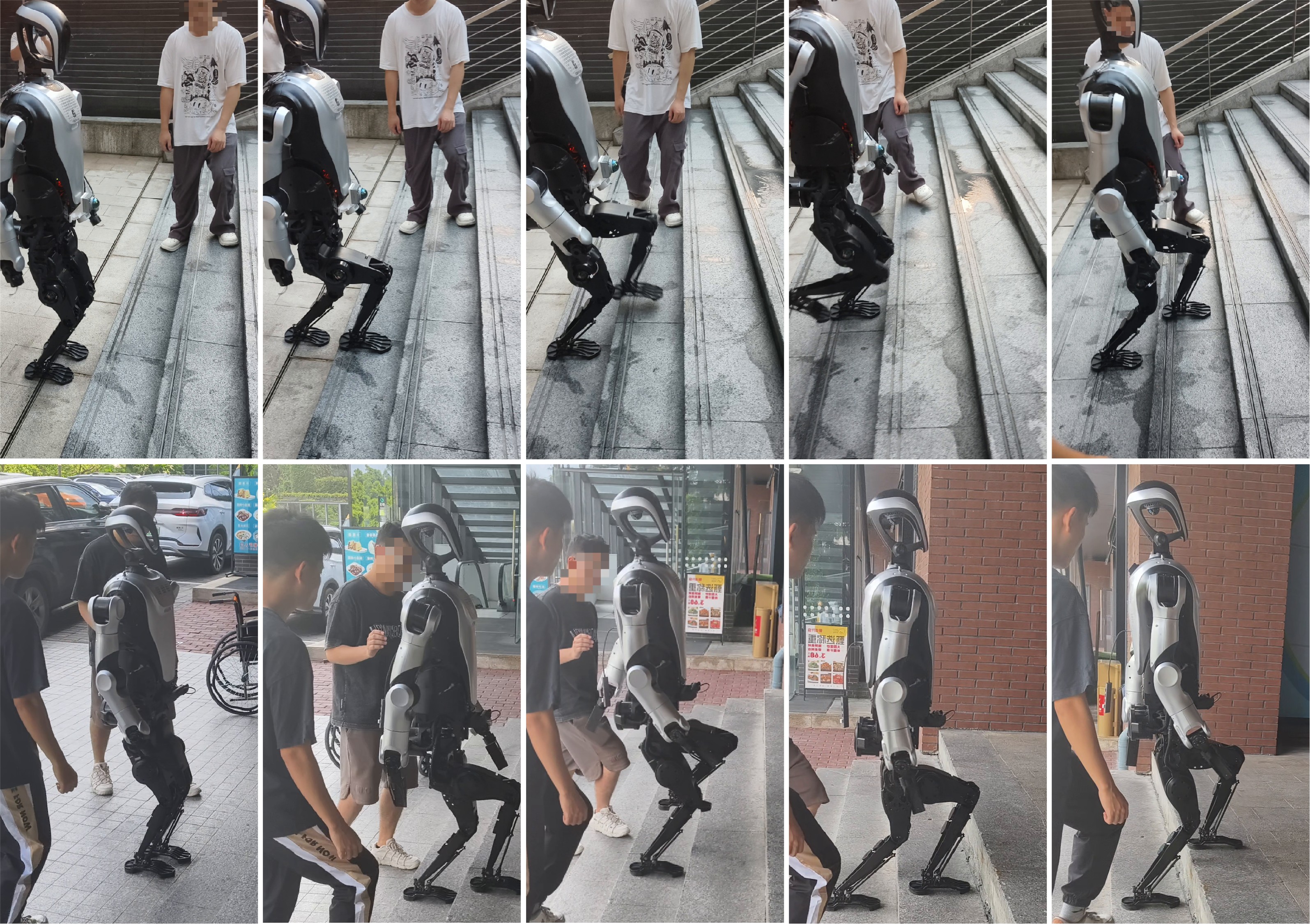}
    \caption{Two outdoor experiments. These snapshots indicate the proposed system is robust in real outdoor environments.}
    \label{fig:outdoor}
\end{figure}

Fig.~\ref{fig:shiwu_2} plots the robot motion trajectories and the feet tracking performance of Fig.~\ref {fig:shiwu_1}'s experiment. The left plot shows the 3D trajectories of the robot base and feet. It can be observed that the robot successfully climbs four consecutive steps, with the foot trajectories highly consistent with the planned foothold regions. The trajectories are smooth and exhibit no noticeable drift, indicating that the closed-loop foothold planning and execution are reliable and effective.  
The right plot compares the desired and actual foot trajectories in the $x$- and $z$-directions, along with the corresponding tracking error curves. 
These results indicate the foot trajectories closely follow the planned curves, with the maximum tracking error in both directions remaining within 40\,mm. The error curves exhibit a periodic pattern, with the biggest errors appearing just before the initiation of each footstep trajectory, reflecting the transition from a stationary to a moving state. Subsequent error peaks correspond to the swing phases of the feet, indicating that the controller can promptly correct trajectory deviations during the swing motion. 

Fig.~\ref{fig:outdoor} presents two outdoor experiments corresponding to Table~\ref{tab:shiwu_results}. In both scenarios, the robot successfully climbed 5–6 steps. In the lower images representing a standard stair, the robot smoothly completed the climb and reached the top. The upper part shows a stair with a protruding edge, where the combination of the protruding edge and the relatively large foothold error of the SS gait finally leads the robot’s toe to collide with the edge.
There is still a failure rate during outdoor long staircase climbing cases, especially in the Single-Step pattern. The reasons may stem from inaccurate state estimation, outdoor lighting conditions or some actuator problem under the single-step pattern.

Overall, the errors are minor and quickly converge, demonstrating that the closed-loop error from perception to execution is effectively suppressed. These findings verify that the proposed perception-planning-control loop not only accomplishes the stair-climbing task but also ensures high-precision foot trajectory tracking and stable locomotion on uneven surfaces.

\section{Conclusions}
\label{char5}  
\renewcommand\arraystretch{2}
This paper presents a vision-guided stair climbing framework for full-size humanoid robots. 
The perception module combines LIO, RGB-D point clouds and proprioceptive observations to extract planes, then outputs stable and low-drift state estimation. 
The planning module selects foothold regions and generates foot trajectories that satisfy stability and safety constraints. 
Both simulation and real-world experiments demonstrate that the proposed method exhibits strong robustness and achieves real-time performance of 20-30 Hz, enabling stable continuous stair climbing in real scenes.
\bibliographystyle{IEEEtran.bst}
\bibliography{IEEEabrv,bib}
\end{CJK}
\end{document}